\documentclass{article}
\usepackage{spconf,amsmath,graphicx}
\usepackage{multirow}
\usepackage{booktabs}
\usepackage{multicol}
\usepackage{url}     
\usepackage{float}
\usepackage{rotating} 

\title{An Experimental Study of the Impact of Pre-Training on the Pruning of a Convolutional Neural Network}

\name{Nathan Hubens $^{\star \dagger}$ \thanks{This research has been conducted in the context of a joint-PhD between the two institutions.} \quad Matei Mancas $^{\star}$ \quad  Bernard Gosselin $^{\star}$  \quad Marius Preda $^{\dagger}$  \quad Titus Zaharia $^{\dagger}$ }
  
\address{$^{\star}$ ISIA Lab, University of Mons \\
      $^{\dagger}$ Artemis, IP Paris}

\begin{document}
\maketitle
\begin{abstract}
In recent years, deep neural networks have known a wide success in various application domains. However, they require important computational and memory resources, which severely hinders their deployment, notably on mobile devices or for real-time applications. Neural networks usually involve a large number of parameters, which correspond to the weights of the network. Such parameters, obtained with the help of a training process, are determinant for the performance of the network. However, they are also highly redundant. The pruning methods notably attempt to reduce the size of the parameter set, by identifying and removing the irrelevant weights. In this paper, we examine the impact of the training strategy on the pruning efficiency. Two training modalities are considered and compared: (1) fine-tuned and (2) from scratch. The experimental results obtained on four datasets (CIFAR10, CIFAR100, SVHN and Caltech101) and for two different CNNs (VGG16 and MobileNet) demonstrate that a network that has been pre-trained on a large corpus (e.g. ImageNet) and then fine-tuned on a particular dataset can be pruned much more efficiently (up to 80\% of parameter reduction) than the same network trained from scratch.

\end{abstract}


\maketitle


\section{Introduction}
Over the last years, Convolutional Neural Networks (CNNs) have exhibited state-of-the-art performance in various computer vision tasks, including image classification and object detection \cite{sota1, sota2}. While their performance continuously improved, such networks kept growing deeper in complexity and depth. This led to increasing needs of parameters storage and convolution operations, inhibiting the use of deep neural networks for applications with memory or processing time limitations. \\

As the majority of the parameters of CNNs are located in the fully-connected layers, recent works have attempted to reduce the memory demands of such networks by removing the fully-connected part of the network, which is replaced by an average pooling layer \cite{no_fc_1, no_fc_2}. Such approaches make it possible to decrease significantly the number of parameters without affecting the accuracy. On the other hand, the convolution layers are those where most of the computations are realized. This computation cost can be reduced by downsampling the feature maps earlier in the network \cite{ds} through various pooling mechanisms or by replacing the convolution operations by factorized convolutions \cite{mobilenet, xception}. \\

Nowadays, a widely used approach when training neural networks for specific applications concerns the fine-tuning \cite{FT}. It consists of using the weights of a model pre-trained on a generic database as initialization and then training the model on a particular dataset. Neural network fine-tuning offers the advantage of making possible to obtain results for particular applications, where the training datasets are not sufficiently large. The dataset used for the pre-training is often very large and contains various examples (e.g. ImageNet \cite{imagenet}), making the network able to extract a large variety of different features. Generally, the network that has been pre-trained on the large dataset also has a very high capacity and can be over-parameterized for the particular dataset. This can lead to redundant or irrelevant weights. Removing such weights will then decrease the number of parameters of the network without significantly degrading its accuracy. \\

In this work, we propose an experimental analysis of the sensitivity of the pruning methods with respect to the training strategy. We consider for evaluation: (1) a network that has been pre-trained on a generic dataset (i.e., ImageNet), and then fine-tuned on a target dataset; and (2) a network that has been directly trained on the target dataset from randomly initialized weights. The experimental results obtained on various datasets (CIFAR10, CIFAR100, SVHN and Caltech101) and for two different networks (VGG16 and MobileNet) demonstrate the superiority of the first approach, which consists of using pre-training process. \\

The rest of the paper is organized as follows. Section 2 briefly presents an overview of the state-of-the-art pruning methods. Section 3 introduces the retained evaluation methodology, with adopted pruning technique, datasets and network architectures. The experimental results obtained are presented and discussed in Section 4. Finally, Section 5 concludes the paper and opens perspectives of future work.


\section{Related Work}

Recently, there has been a line of work concerning the lottery ticket hypothesis \cite{lottery, lottery2, lottery3}. This hypothesis argues that, in a regular neural network architecture, there exists a sub-network that can be trained to the same level of performance as the original one, as long as it starts from the original initial conditions. This means that a neural network does not require all of its parameters to perform correctly and that having an over-parametrized neural network is only useful to find the "winning ticket". In practice, those sub-networks can be found by training the original network to convergence, removing the unnecessary weights, then resetting the value of the remaining weights to their original value. \\

Early work on pruning methods dates back to Optimal Brain Damage \cite{brain_damage} and Optimal Brain Surgeon \cite{brain_surgeon}, where the weights are pruned based on the Hessian of the loss function. More recently, \cite{han} proposes to prune the weights with small magnitude. This kind of pruning, performed on individual weights, is called unstructured pruning \cite{struct}, as there is no intent to preserve any structure or geometry in the network. The unstructured pruning methods are the most efficient as they prune weights at the most fine-grained level. However, they lead to sparse weight matrices. Consequently, taking the advantage of the pruning results requires dedicated hardware or libraries able to efficiently deal with such sparsity. \\

To overcome this limitation, so-called structured pruning methods have been introduced. In this case, the pruning is operated at the level of the filters, kernels or vectors \cite{recent, channel_pruning}. One of the most popular structured pruning technique is the so-called filter pruning. Here, at each pruning stage, complete convolution filters are removed and the convolution structure remains unchanged. The filter selection methods can be based on the filter weight norm \cite{li}, average percentage of zeros in the output \cite{apoz}, or on the influence of each channel on the final loss \cite{molch}. Intrinsic correlation within each layer can also be exploited to remove redundant channels \cite{suau}. Pruning can also be combined with other compression techniques such as weight quantization \cite{xnor}, low-rank approximations of weights \cite{denton}, knowledge distillation \cite{KD, KD2} to further reduce the size of the network. \\

In our work, we have adopted a structured pruning approach, performed at the filter level. As in \cite{li, Liu}, we use the $l_1$-norm to determine the importance of filters and select those to prune. We decided to adopt an iterative pruning process, where the pruning is performed one layer at a time, and the network is retrained after each pruning phase, to allow it to recover from the loss of some parameters. 


\section{Methodology}

In this section, we describe in detail the methodology we followed for training and pruning a model.

\subsection{Pruning method}
\label{pruning_method}
\begin{figure}[!htpb]
\begin{minipage}[b]{.49\linewidth}
  \centering
  \centerline{\includegraphics[width=3.62cm]{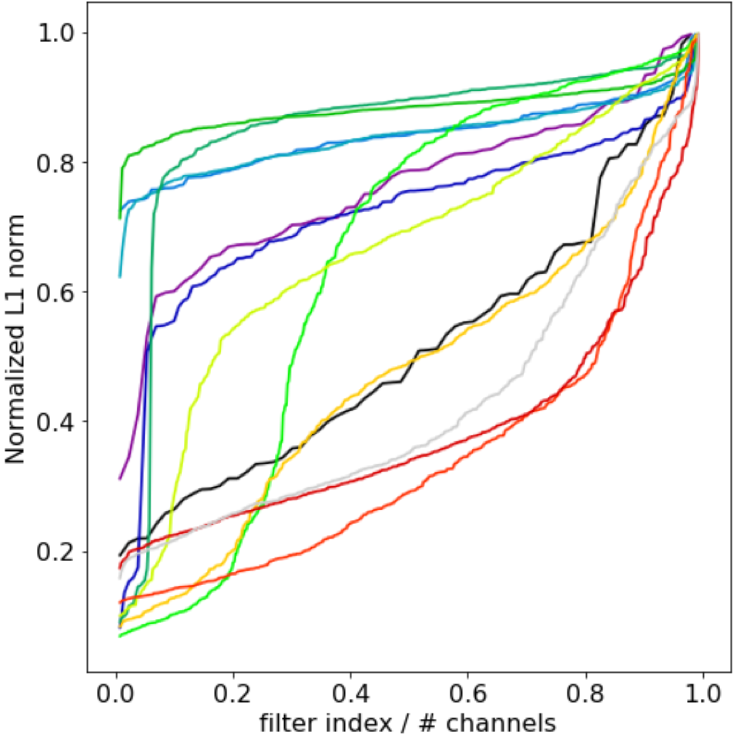}}
  {(a) Filters ranked by ascending $l_1$-norm for VGG16 trained on MNIST.}
\end{minipage}
\begin{minipage}[b]{0.49\linewidth}
  \centering
  \centerline{\includegraphics[width=4.47cm]{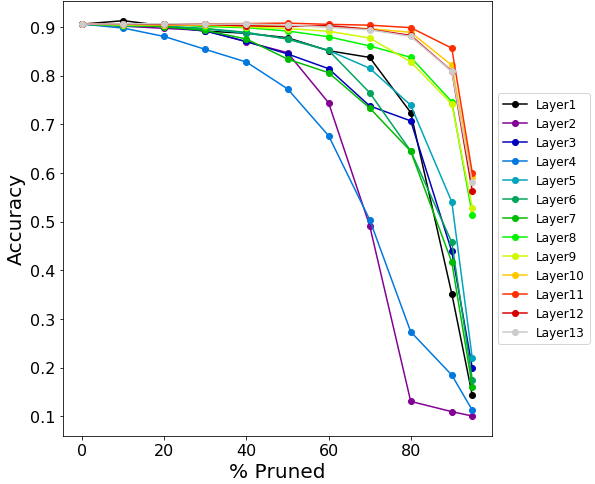}}
   {(b) Sensitivity of each convolutional layer pruned individually.}
   \end{minipage}
   \caption{Visualization of the sensitivity to pruning of a VGG-16 trained on MNIST.}
\label{sensitivity}
\end{figure}

The pruning method \cite{li} used in our work prunes the filters of a trained network that have lowest sensitivity to pruning. The sensitivity to pruning of a layer is closely related to the $l_1$-norm of its filters. Figure \ref{sensitivity}b shows the sensitivity to pruning of each layer of a VGG16 network fine-tuned on the MNIST dataset \cite{mnist}. The layers that are the most sensitive, i.e. the layers where the accuracy drops the fastest when removing filters, are also the layers with the most high $l_1$-norm filters, as illustrated in Figure \ref{sensitivity}a. \\

Based on this observation, we propose to remove filters with the lowest $l_1$-norm as they produce feature maps with weaker activations compared to other filters in that layer. 

\begin{figure}[!htpb]
  \centering
  \includegraphics[width=\linewidth]{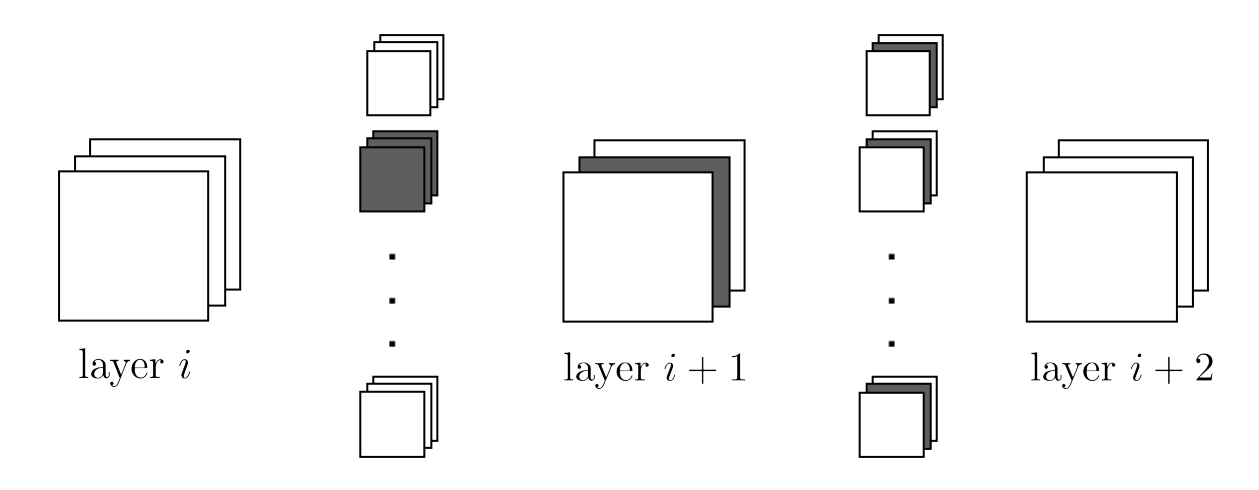}
  \caption{Representation of filter pruning of a convolutional layer. The pruned filter, as well as its corresponding feature map and kernels in the following layer, are removed.}
  \label{pruning}
\end{figure}

More precisely, the procedure of pruning $m$ filters from a layer $i$ is the following: 

\begin{enumerate}
  \setlength{\itemsep}{1pt}
  \setlength{\parskip}{0pt}
  \setlength{\parsep}{0pt}
	\item  Compute the $l_1$-norm of each filter of layer $i$.
	\item Sort the filters by their $l_1$-norm.
	\item Prune $m$ filters with the smallest $l_1$-norm and their corresponding feature maps in layer $i+1$. The removed feature maps will not participate to the next computations, thus their corresponding kernels in the successive convolution filters are also removed (Figure \ref{pruning}).
	\item Retrain the whole network until new convergence.
\end{enumerate}

\subsection{Datasets}

We have carried out our experiments on four different datasets, with a good variability of number of classes, image resolution and content. As targeted datasets, we have CIFAR-10 and CIFAR-100 \cite{cifar}, consisting of RGB images of $32 \times 32$ pixels, labelled respectively over 10 and 100 classes. We also retained SVHN \cite{svhn}, a real-world image dataset for recognizing digits and numbers in natural scene images of size $32 \times 32$. Finally, we have considered the Caltech101 corpus \cite{caltech}, consisting of pictures of objects belonging to 101 categories, of size of approximately $300 \times 200$ pixels. \\

\subsection{Network architectures}

In order to validate our results, we have adopted two well-known network architectures. The first one is VGG16 \cite{sota1}. Here, we have replaced the original fully-connected layers by a Global Average Pooling layer and 2 narrow fully-connected layers. In this way, most parameters are contained in the convolutional layers. The network thus consists of 13 convolutional layers and 2 fully-connected layers. \\

The second network retained is MobileNetV1 \cite{mobilenet}, specifically designed to achieve efficiency both in parameter number and in computation complexity. MobileNet uses a factorized form of convolutions called Depthwise Separable Convolutions. The MobileNet architecture used in our experiments thus consists of one standard convolution layer acting on the input image, 13 depthwise separable convolutions, and finally a global average pooling and 2 fully connected layers. \\

Whatever the network retained, in experiments, Network-A will refer to the network pre-trained on ImageNet and fine-tuned on the target dataset, while Network-B refers to the same network trained on the target datasets from scratch.

\section{Experimental results}

In this section, we compare the pruning efficiency of both fine-tuned and trained from scratch networks. Our models are first trained until convergence, using the Adam optimizer, with an initial learning rate of $0.001$ and a step decay scheduling, reducing the learning rate by a factor $10$ every $40$ epochs. They are then tested on the validation set to get the baseline accuracy. \\

After each pruning phase, a retraining is performed during $5$ epochs, with the lowest learning rate reached during baseline training. We also monitor the accuracy on the validation set, and proceed to another pruning phase if it has not dropped by more than $1\%$ from the baseline accuracy. 


\subsection{VGG16}

\begin{figure}[!htpb]

\begin{minipage}[b]{.49\linewidth}
  \centering
  \centerline{\includegraphics[width=3.66cm]{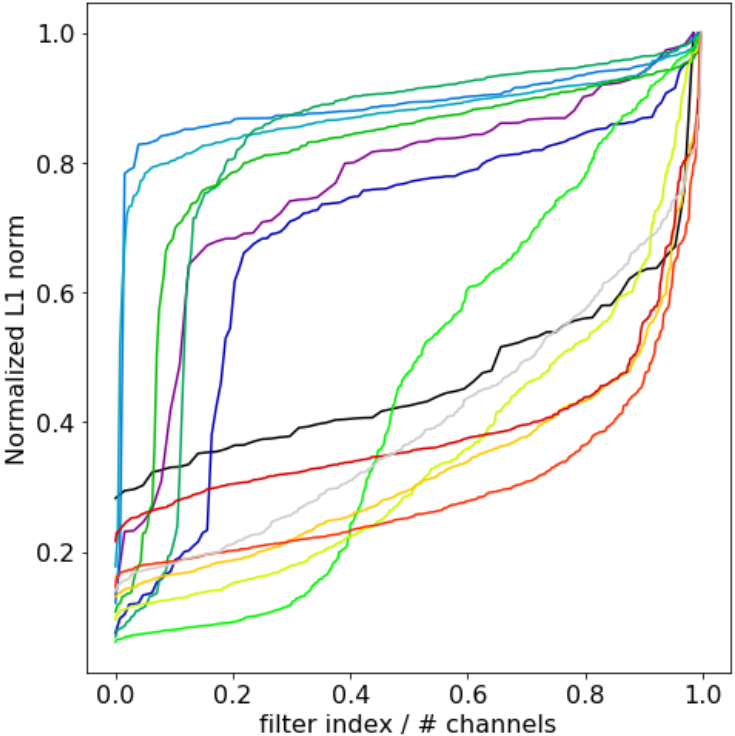}}
  {(a) Filters ranked by ascending $l_1$-norm for Network-A}
\end{minipage}
\hfill
\begin{minipage}[b]{0.49\linewidth}
  \centering
  \centerline{\includegraphics[width=4.44cm]{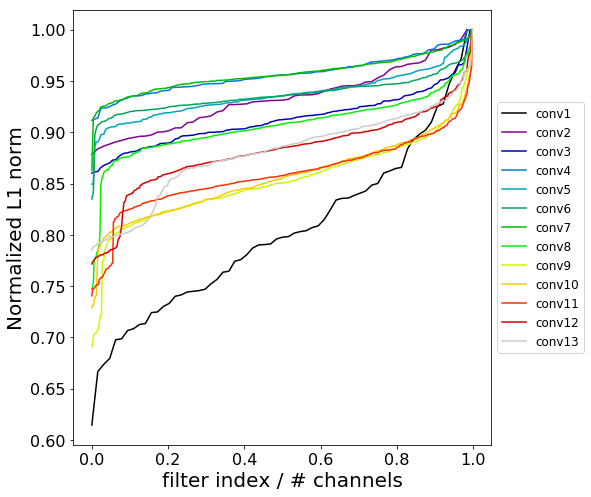}}
   {(b) Filters ranked by ascending $l_1$-norm for Network-B}
   \end{minipage}

\caption{Visualization of the importance of filters of VGG-16 trained on CIFAR-10. Filters are ranked by $l_1$-norm. }
\label{cifar10_norm}
\end{figure}

To select the layer that will be pruned and decide of many filters to be removed, we conduct a sensitivity analysis on the network. As shown in Figure \ref{cifar10_norm}a, most of the low-norm filters are contained in the later layers. This suggests that those layers will be less sensitive to pruning than the others. \\

On the other hand, Figure \ref{cifar10_norm}b shows that Network-B has a more even repartition of filter norms, which suggests that it is more sensitive to pruning than Network-A, specifically in the later layers.

This observation is confirmed by the results of pruning, summarized in Table \ref{layer_pruning_vgg}, which shows that indeed, most of the pruning of Network-A can be performed in the later layers while pruning of Network-B is more evenly distributed. \\

\begin{table}[!htpb]
  \begin{center}
    \caption{Parameters remaining for each layer after the pruning of VGG16, trained on CIFAR10.}
   \label{layer_pruning_vgg}
    \small
    \begin{tabular}{cc|c|c}
      \textbf{Layer type} & \#Params & Network-A & Network-B  \\
      \hline
     Conv 1 & 1792 & 1792 & 1792 \\
     Conv 2 & 36,928 & 36,928 & 36,928 \\
     Conv 3 & 73,856 & 73,856 & 73,856 \\
     Conv 4 & 147,584 & 147,584 & 129,136 \\
     Conv 5 & 295,168 & 295,168 & 161,440\\
     Conv 6 & 590,080 &  590,080 & 230,560 \\
     Conv 7 & 590,080 &  590,080 & 230,560  \\
     Conv 8 & 1,180,160 & 442,560 & 553,344  \\
     Conv 9 & 2,359,808 & 331,968 & 1,327,488 \\
     Conv 10 & 2,359,808 & 331,968 & 1,327,488 \\
     Conv 11 & 2,359,808 & 221,312 & 1,327,488 \\
     Conv 12 & 2,359,808 & 147,584 & 1,327,488 \\
     Conv 13 & 2,359,808 & 147,584  & 1,327,488 \\
     Linear & 262,656 & 66,048 & 197,120  \\
     Linear & 5130 & 5130 & 5130 \\
      \hline
      \textbf{Total} & 14.98M & 3.43M & 8.26M \\
      \hline 
    \end{tabular}
  \end{center}      
\end{table}

Table \ref{vgg_res} summarizes the results on all the tested datasets, and shows the resulting number of parameters, their corresponding storage size, as well as the number of floating point operation (FLOPs) needed for an input image to traverse the whole network at testing phase. We can observe that for all the tested datasets, pruning is more effective in terms of parameters removed for Network-A than for Network-B.
\\

A second observation is that a lower number of parameters do not necessarily lead to a reduction in FLOPs. This phenomenon can be explained by the fact that, while most of the parameters are contained in the later layers, most of the operations are performed in the first ones, where the resolutions of the activation maps are higher. For Network-A, most of the pruning is performed in the later layers. In contrast, for Network-B, the pruning is more distributed throughout the network. Thus, Network-A  has the fewest parameters but Network-B often has the fewest FLOPs.

\begin{table}[!htpb]
  \begin{center}
    \caption{Results of the pruning on VGG16 for the four studied datasets.}
    \label{vgg_res}
    \setlength\tabcolsep{4.5pt}
    \small
    \begin{tabular}{c|lccc}
    \toprule
     Dataset & Network & Params (M) & FLOP (M) & Size (MB) \\
      \hline
 {{\textbf{CIFAR10}}} & Baseline & 14.98 & 627.48 & 57.22 \\
      & A-pruned & \textbf{3.43} & 421.20 & \textbf{13.15}  \\
      & B-pruned & 8.26 & \textbf{397.85} & 31.57  \\      
      \hline
  {{\textbf{CIFAR100}}} & Baseline & 15.03 & 627.57 & 57.39  \\
      & A-pruned  & \textbf{8.26} & 503.47 & \textbf{31.57}   \\
      & B-pruned & 8.69 & \textbf{444.56} & 33.24  \\      
      \hline
  {{\textbf{SVHN}}} & Baseline & 14.98 & 627.48 & 57.22  \\
      & A-pruned  & \textbf{3.15} & 414.12 & \textbf{12.10}   \\
      & B-pruned & 4.08 & \textbf{311.04} & 15.61  \\      
      \hline
  {{\textbf{Caltech101}}} & Baseline & 15.03 & 30,720.99 & 57.40  \\
      & A-pruned  & \textbf{8.44} & \textbf{25,402.11} & \textbf{32.28}   \\
      & B-pruned & 13.59 & 30,171.17 & 51.93  \\  
    \bottomrule  
    \end{tabular}
  \end{center}      
\end{table}

\begin{figure}[!htpb]
\begin{minipage}[b]{.49\linewidth}
  \centering
  \centerline{\includegraphics[width=3.7cm]{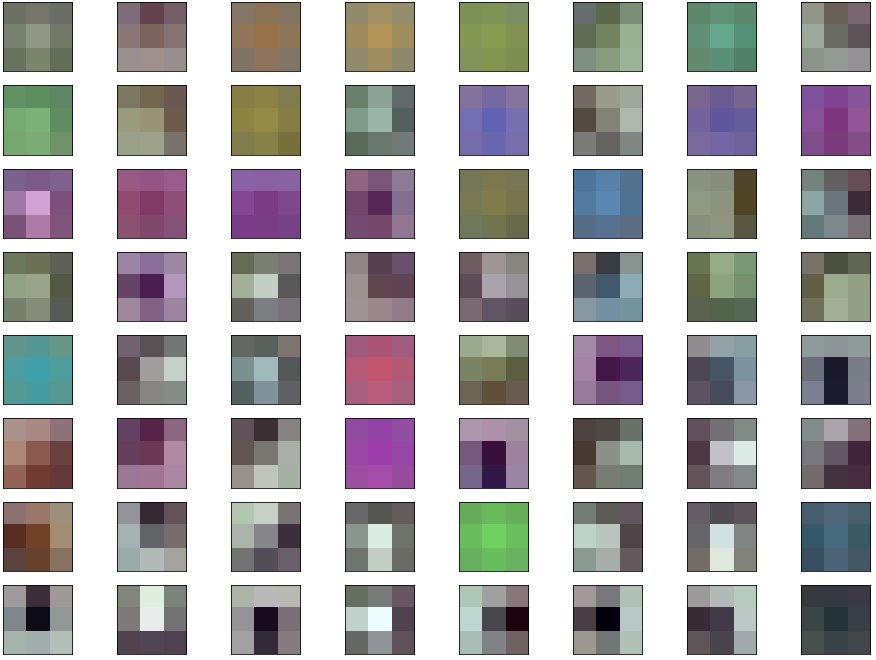}}
  {(a) Filters ranked by ascending $l_1$-norm for Network-A}
\end{minipage}
\hfill
\begin{minipage}[b]{0.49\linewidth}
  \centering
  \centerline{\includegraphics[width=3.7cm]{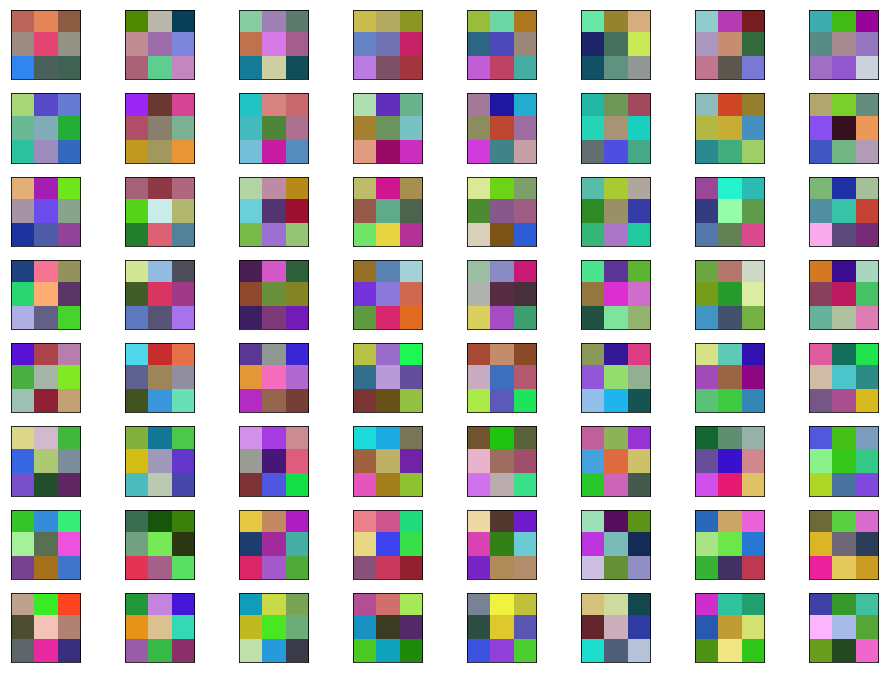}}
   {(b) Filters ranked by ascending $l_1$-norm for Network-B}
   \end{minipage}

\caption{Visualization of the 64 filters in the first convolutional layer of VGG-16 trained on CIFAR-10. Network-A filters have more structure than those of Network-B.}
\label{cifar10_filters}
\end{figure}

A closer look at the filters of the first convolutional layer (Figure \ref{cifar10_filters}) exhibits the difference of learned filters. The reason of this difference is because the pre-training of Network-A helped to find useful filters on the ImageNet database. As Network-B only had access to fewer and smaller images, it could not learn the same kind of filters by itself and had to distribute the feature extraction across the network, reason why the later layers are more sensitive to pruning.


\subsection{MobileNet}

The Depthwise Separable Convolutions operations that are used in MobileNet are composed of two operations. The first is a Depthwise Convolution, that filters each input map independently. The second operation is called a Pointwise Convolution, that combines the results of the previous operation. The Pointwise Convolution is similar to a regular convolution but using filter dimensions of $1 \times 1$. As most of the parameters are contained in the second operation, we decided to operate the pruning only on Pointwise Convolutions. 

\begin{figure}[!htpb]
\begin{minipage}[b]{.49\linewidth}
  \centering
  \centerline{\includegraphics[width=3.75cm]{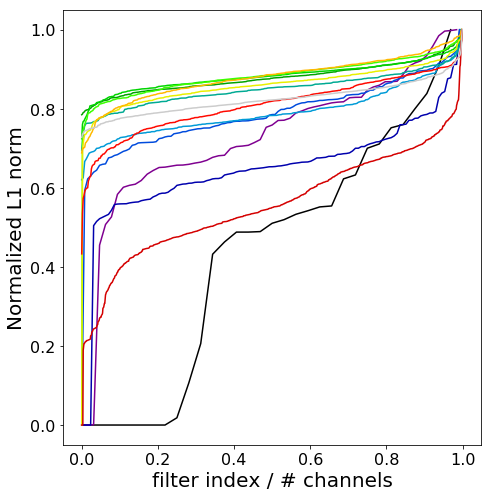}}
  {(a) Filters ranked by ascending $l_1$-norm for Network-A}
\end{minipage}
\hfill
\begin{minipage}[b]{0.49\linewidth}
  \centering
  \centerline{\includegraphics[width=4.42cm]{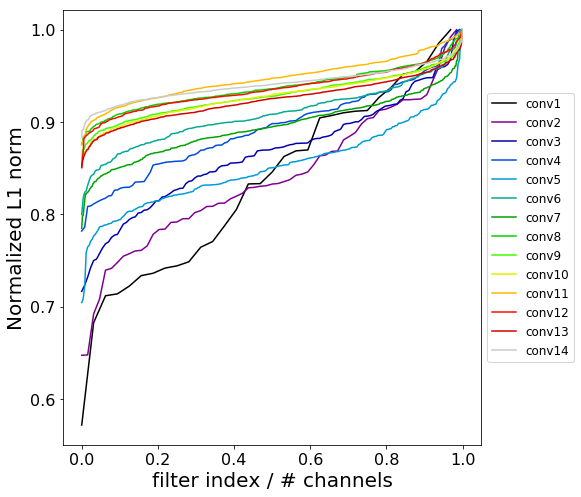}}
  {(b) Filters ranked by ascending $l_1$-norm for Network-B}
  \end{minipage}

\caption{Visualization of the importance of filters of MobileNet trained on CIFAR-10. Filters are ranked by $l_1$-norm.}
\label{cifar100_norm}
\end{figure}

We follow the same process as for VGG16 and perform a sensitivity analysis on MobileNet before deciding which layer and how many filters to remove. The difference between the sensitivity of Network-A (Figure \ref{cifar100_norm}a) and Network-B (Figure \ref{cifar100_norm}b) is less clear than in the case of VGG16 but some useful information can still be extracted. The highest norm filters of Network-B are still in the later layers, while it isn't necessarily the case for Network-A. This again suggests that Network-A can be pruned further in the later layers, and confirmed in Table \ref{layer_pruning_mobilenet}.

\begin{table}[!htpb]
  \begin{center}
   \caption{Parameters remaining for each layer after the pruning of MobileNet, trained on CIFAR10. For clarity, only the first convolution and pointwise convolutions are represented.}
   \label{layer_pruning_mobilenet}
    \small
    \begin{tabular}{cc|c|c}
      \textbf{Layer type} & \#Params & Network-A & Network-B  \\
      \hline
     Conv 1 & 864  & 567  & 864 \\
     Conv 2 & 2048 & 1344 & 2048 \\
     Conv 3 & 8192 & 8192 & 8192 \\
     Conv 4 & 16,384 & 16,384 & 16,384  \\
     Conv 5 & 32,768 & 32,768 & 28,672 \\
     Conv 6 & 65,536 & 65,536 & 50,176 \\
     Conv 7 & 131,072 & 98,304 & 64,512  \\
     Conv 8 & 262,144 & 147,456 & 82,944  \\
     Conv 9 & 262,144 & 147,456 & 82,944 \\
     Conv 10 & 262,144 & 147,456 & 82,944 \\
     Conv 11 & 262,144 & 98,304 & 82,944 \\
     Conv 12 & 262,144 & 65,536 & 78,336 \\
     Conv 13 & 524,288 & 65,536 & 82,688 \\
     Conv 14 & 1,048,576 & 32,768 & 97,280 \\
     Linear & 524,800 & 66,048 & 164,352 \\
     Linear & 5130 & 5130 & 5130 \\
      \hline
      \textbf{Total} & 3.76M & 1.05M & 0.98M \\
      \hline 
    \end{tabular}
  \end{center}      
\end{table}

The results of the pruning of MobileNet on the different datasets are summarized in Table \ref{mobilenet_res}. As it was the case for VGG16, Network-A can also be pruned further than Network-B, even if the difference is smaller in this case.

\begin{figure}[!htpb]
\begin{minipage}[b]{.48\linewidth}
  \centering
  \centerline{\includegraphics[width=3.7cm]{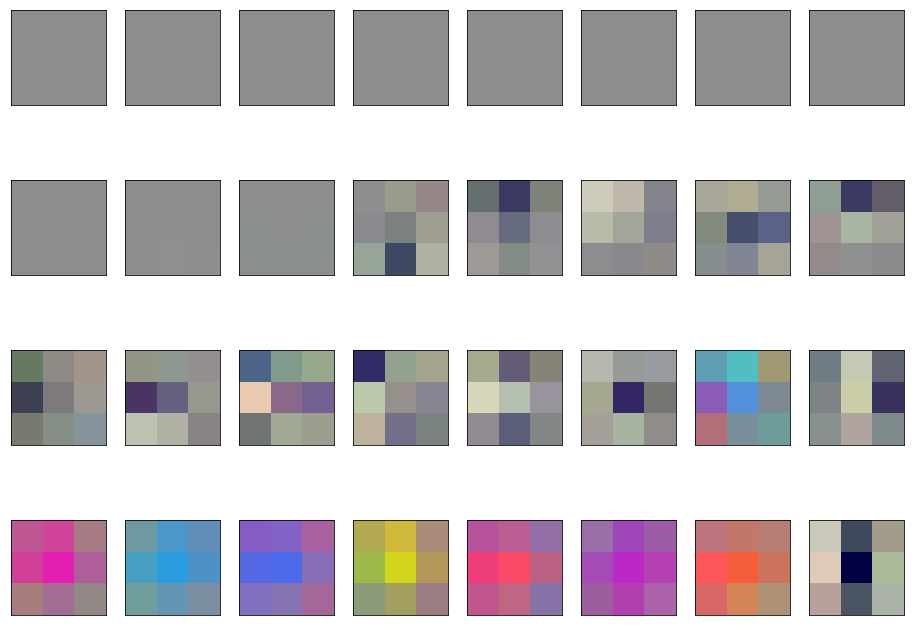}}
  {(a) Filters ranked by ascending $l_1$-norm for Network-A}
\end{minipage}
\hfill
\begin{minipage}[b]{0.48\linewidth}
  \centering
  \centerline{\includegraphics[width=3.7cm]{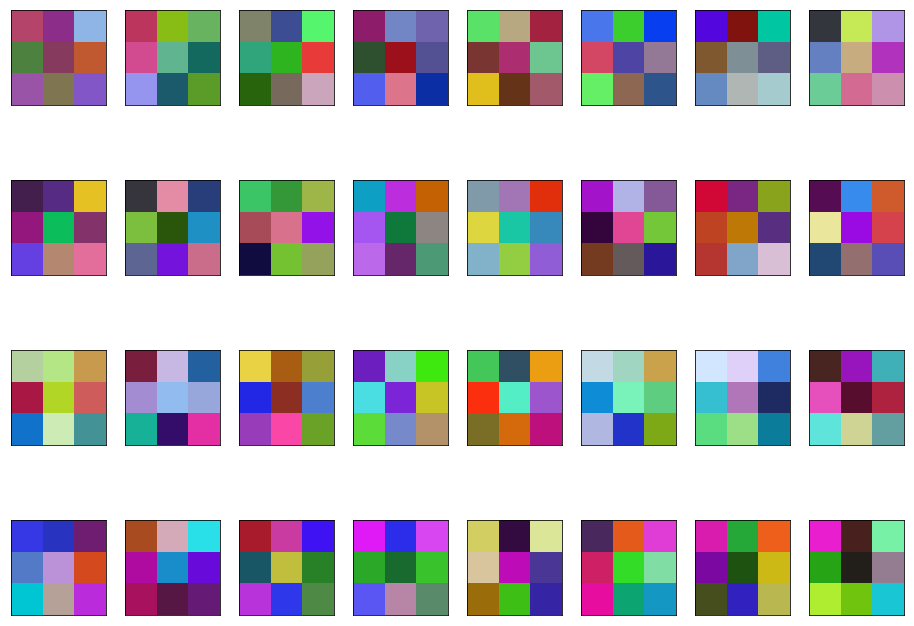}}
   {(b) Filters ranked by ascending $l_1$-norm for Network-B}
   \end{minipage}

\caption{Visualization of the 32 filters in the first convolutional layer of MobileNet trained on CIFAR-10. Filters of Network-A exhibit more structure than the filters of Network-B.}
\label{cifar100_fm}
\end{figure}

\begin{table}[!htpb]
  \begin{center}

    \label{tab:table1}
    \caption{Results of the pruning on MobileNet for the four studied datasets.}
    \label{mobilenet_res}
    \setlength\tabcolsep{4.5pt}
    \small
    \begin{tabular}{c|lccc}
    \toprule
     Dataset & Network & Params (M) & FLOP (M) & Size (MB) \\
    \hline
  {{\textbf{CIFAR10}}} & Baseline & 3.76 & 24.23 & 14.58 \\
      & A-pruned & 1.05 & 13.83 & 4.26  \\
      & B-pruned & \textbf{0.98} & \textbf{12.26} & \textbf{3.99}  \\      
      \hline
  {{\textbf{CIFAR100 }}} & Baseline & 3.80 & 24.32 & 14.76  \\
      & A-pruned  & \textbf{2.54} & \textbf{21.42} & \textbf{9.94}   \\
      & B-pruned & 3.12 & 21.89 & 12.14  \\      
      \hline
  {{\textbf{SVHN}}} & Baseline & 3.76 & 24.23 & 14.58  \\
      & A-pruned  & \textbf{1.50} & 17.66 & \textbf{5.99}   \\
      & B-pruned & 1.75 & \textbf{16.57} & 6.94  \\      
      \hline
  {{\textbf{Caltech101}}} & Baseline & 3.81 & 1136.59 & 14.76  \\
      & A-pruned  & \textbf{3.08} & \textbf{1078.30} & \textbf{11.99}   \\
      & B-pruned & 3.43 & 1105.83 & 13.32  \\
      \bottomrule
    \end{tabular}
  \end{center}      
\end{table}

Again, looking at the filters of the first convolutional layer exhibits the difference of learned filters between Network-A (Figure \ref{cifar100_fm}a) and Network-B (Figure \ref{cifar100_fm}b). The filters of Network-A show some structure where those from Network-B don't appear to. Moreover, as it was suggested in Figure \ref{cifar100_norm}a, Network-A has some really low-norm filters, that don't seem to extract useful information, and can thus be removed.


\section{Conclusion and Perspectives}

In this paper, we have investigated the impact of the training process on the number of relevant filters involved in a CNN. In particular, we have compared the sensitivity of filter pruning of a network that has been fine-tuned to the sensitivity of a network trained from scratch. \\

Experiments have been conducted on four different datasets (CIFAR10, CIFAR100, SVHN, Caltech101) and for two different network architectures (VGG16, MobileNet). Results have shown that a CNN that has been pre-trained and then fine-tuned on a target dataset is less sensitive to pruning and thus, can be pruned further than the same network trained from-scratch on the target dataset. \\

This also supposes that the methodology of training has a strong impact on the part of the network where features are extracted. Pre-training helps the network to be discriminant in early layers and thus allows for more pruning in later layers, where most of the parameters are contained. On the other hand, training a network from randomly initialized weights makes its layers more evenly discriminant and thus the pruning has to be more distributed across the layers, resulting in fewer parameters removed but often a bigger reduction in computation operations. \\ 

Further research can be conducted on using this combination of sensitivity analysis and pruning to improve existing architectures of CNNs or find new ones that are more efficient. It can also lead to finding new training strategies either targeting a low number of parameters after pruning, as it is the case with fine-tuning, or targeting fewer computations, as it is the case for a network trained from-scratch. \\





\bibliographystyle{IEEEbib}
\bibliography{biblio}

\begin{thebibliography}{10}

\bibitem{sota1}
Karen Simonyan and Andrew Zisserman,
\newblock ``Very deep convolutional networks for large-scale image
  recognition,''
\newblock in {\em 3rd International Conference on Learning Representations,
  {ICLR} 2015, San Diego, CA, USA, May 7-9, 2015, Conference Track
  Proceedings}, 2015.

\bibitem{sota2}
Christian Szegedy, Wei Liu, Yangqing Jia, Pierre Sermanet, Scott Reed, Dragomir
  Anguelov, Dumitru Erhan, Vincent Vanhoucke, and Andrew Rabinovich,
\newblock ``Going deeper with convolutions,''
\newblock in {\em 2015 {IEEE} {Conference} on {Computer} {Vision} and {Pattern}
  {Recognition} ({CVPR})}, June 2015, pp. 1--9,
\newblock ISSN: 1063-6919, 1063-6919.

\bibitem{no_fc_1}
Christian Szegedy, Vincent Vanhoucke, Sergey Ioffe, Jonathon Shlens, and
  Zbigniew Wojna,
\newblock ``Rethinking the inception architecture for computer vision,''
\newblock in {\em Conference on Computer Vision and Pattern Recognition,
  {CVPR}}, 2016.

\bibitem{no_fc_2}
Min Lin, Qiang Chen, and Shuicheng Yan,
\newblock ``Network in network,''
\newblock in {\em 2nd International Conference on Learning Representations,
  {ICLR} 2014, Banff, AB, Canada, April 14-16, 2014, Conference Track
  Proceedings}, 2014.

\bibitem{ds}
Kaiming He and Jian Sun,
\newblock ``Convolutional neural networks at constrained time cost,''
\newblock {\em CoRR}, vol. abs/1412.1710, 2014.

\bibitem{mobilenet}
Andrew~G. Howard, Menglong Zhu, Bo~Chen, Dmitry Kalenichenko, Weijun Wang,
  Tobias Weyand, Marco Andreetto, and Hartwig Adam,
\newblock ``Mobilenets: Efficient convolutional neural networks for mobile
  vision applications,''
\newblock {\em CoRR}, vol. abs/1704.04861, 2017.

\bibitem{xception}
Fran{\c{c}}ois Chollet,
\newblock ``Xception: Deep learning with depthwise separable convolutions,''
\newblock {\em CoRR}, vol. abs/1610.02357, 2016.

\bibitem{FT}
Jason Yosinski, Jeff Clune, Yoshua Bengio, and Hod Lipson,
\newblock ``How transferable are features in deep neural networks?,''
\newblock {\em CoRR}, vol. abs/1411.1792, 2014.

\bibitem{imagenet}
Jia Deng, Wei Dong, Richard Socher, Li-Jia Li, Kai Li, and Li~Fei-Fei,
\newblock ``{ImageNet}: A large-scale hierarchical image database,''
\newblock in {\em 2009 {IEEE} Conference on Computer Vision and Pattern
  Recognition}, 2009, pp. 248--255.

\bibitem{lottery}
Jonathan Frankle and Michael Carbin,
\newblock ``The lottery ticket hypothesis: Training pruned neural networks,''
\newblock {\em CoRR}, vol. abs/1803.03635, 2018.

\bibitem{lottery2}
Jonathan Frankle, Gintare~Karolina Dziugaite, Daniel~M. Roy, and Michael
  Carbin,
\newblock ``The lottery ticket hypothesis at scale,''
\newblock {\em CoRR}, vol. abs/1903.01611, 2019.

\bibitem{lottery3}
Ari~S. Morcos, Haonan Yu, Michela Paganini, and Yuandong Tian,
\newblock ``One ticket to win them all: generalizing lottery ticket
  initializations across datasets and optimizers,''
\newblock {\em ArXiv}, vol. abs/1906.02773, 2019.

\bibitem{brain_damage}
Yann LeCun, John~S. Denker, and Sara~A. Solla,
\newblock ``Optimal brain damage,''
\newblock in {\em Advances in Neural Information Processing Systems 2}, D.~S.
  Touretzky, Ed., pp. 598--605. Morgan-Kaufmann, 1990.

\bibitem{brain_surgeon}
Babak Hassibi, David~G. Stork, Gregory Wolff, and Takahiro Watanabe,
\newblock ``Optimal brain surgeon: Extensions and performance comparisons,''
\newblock in {\em Advances in Neural Information Processing Systems 6}, pp.
  263--270. Morgan-Kaufmann, 1994.

\bibitem{han}
Song Han, Jeff Pool, John Tran, and William~J. Dally,
\newblock ``Learning both weights and connections for efficient neural
  networks,''
\newblock {\em CoRR}, vol. abs/1506.02626, 2015.

\bibitem{struct}
Sajid Anwar, Kyuyeon Hwang, and Wonyong Sung,
\newblock ``Structured pruning of deep convolutional neural networks,''
\newblock {\em CoRR}, vol. abs/1512.08571, 2015.

\bibitem{recent}
Jian Cheng, Peisong Wang, Gang Li, Qinghao Hu, and Hanqing Lu,
\newblock ``Recent advances in efficient computation of deep convolutional
  neural networks,''
\newblock {\em CoRR}, vol. abs/1802.00939, 2018.

\bibitem{channel_pruning}
Yihui He, Xiangyu Zhang, and Jian Sun,
\newblock ``Channel pruning for accelerating very deep neural networks,''
\newblock {\em CoRR}, vol. abs/1707.06168, 2017.

\bibitem{li}
Hao Li, Asim Kadav, Igor Durdanovic, Hanan Samet, and Hans~Peter Graf,
\newblock ``Pruning filters for efficient convnets,''
\newblock {\em CoRR}, vol. abs/1608.08710, 2016.

\bibitem{apoz}
Hengyuan Hu, Rui Peng, Yu{-}Wing Tai, and Chi{-}Keung Tang,
\newblock ``Network trimming: {A} data-driven neuron pruning approach towards
  efficient deep architectures,''
\newblock {\em CoRR}, vol. abs/1607.03250, 2016.

\bibitem{molch}
Pavlo Molchanov, Stephen Tyree, Tero Karras, Timo Aila, and Jan Kautz,
\newblock ``Pruning convolutional neural networks for resource efficient
  transfer learning,''
\newblock {\em CoRR}, vol. abs/1611.06440, 2016.

\bibitem{suau}
Xavier Suau, Luca Zappella, Vinay Palakkode, and Nicholas Apostoloff,
\newblock ``Principal filter analysis for guided network compression,''
\newblock {\em CoRR}, vol. abs/1807.10585, 2018.

\bibitem{xnor}
Mohammad Rastegari, Vicente Ordonez, Joseph Redmon, and Ali Farhadi,
\newblock ``Xnor-net: Imagenet classification using binary convolutional neural
  networks,''
\newblock {\em CoRR}, vol. abs/1603.05279, 2016.

\bibitem{denton}
Emily Denton, Wojciech Zaremba, Joan Bruna, Yann LeCun, and Rob Fergus,
\newblock ``Exploiting linear structure within convolutional networks for
  efficient evaluation,''
\newblock {\em CoRR}, vol. abs/1404.0736, 2014.

\bibitem{KD}
Geoffrey Hinton, Oriol Vinyals, and Jeffrey Dean,
\newblock ``Distilling the knowledge in a neural network,''
\newblock in {\em NIPS Deep Learning and Representation Learning Workshop},
  2015.

\bibitem{KD2}
Jimmy Ba and Rich Caruana,
\newblock ``Do deep nets really need to be deep?,''
\newblock in {\em Advances in Neural Information Processing Systems}, pp.
  2654--2662. Curran Associates, Inc., 2014.

\bibitem{Liu}
Zhuang Liu, Mingjie Sun, Tinghui Zhou, Gao Huang, and Trevor Darrell,
\newblock ``Rethinking the value of network pruning,''
\newblock {\em CoRR}, vol. abs/1810.05270, 2018.

\bibitem{mnist}
Yann LeCun and Corinna Cortes,
\newblock ``{MNIST} handwritten digit database,''
\newblock 2010.

\bibitem{cifar}
Alex Krizhevsky,
\newblock ``Learning multiple layers of features from tiny images,''
\newblock 2009.

\bibitem{svhn}
Yuval Netzer, Tao Wang, Adam Coates, Alessandro Bissacco, Bo~Wu, and Andrew~Y.
  Ng,
\newblock ``Reading digits in natural images with unsupervised feature
  learning,''
\newblock in {\em Workshop on Deep Learning and Unsupervised Feature Learning,
  {NeurIPS}}, 2011.

\bibitem{caltech}
Fei-Fei Li, Rob Fergus, and Pietro Perona,
\newblock ``Learning generative visual models from few training examples: An
  incremental bayesian approach tested on 101 object categories,''
\newblock in {\em 2004 Conference on Computer Vision and Pattern Recognition
  Workshop}, June 2004, pp. 178--178.

\end{thebibliography}
\end{document}